%% file: main.tex
\documentclass{article}

\usepackage{microtype}
\usepackage{graphicx}
\usepackage{subfigure}
\usepackage{booktabs} % for professional tables
\usepackage{hyperref}

\usepackage[accepted]{icml2021}
\usepackage{amsmath}
\usepackage{amssymb}
\usepackage{multirow} % for results_exp2 table
\usepackage{enumitem} % for pretty display of research questions
\newlist{questions}{enumerate}{2}
\setlist[questions,1]{label=RQ\arabic*.,ref=RQ\arabic*}
\setlist[questions,2]{label=(\alph*),ref=\thequestionsi(\alph*)}

\icmltitlerunning{Offline Clustering Approach to Self-supervised Learning for Class-imbalanced Image Data}

\begin{document}

\twocolumn[
\icmltitle{Offline~Clustering~Approach~to~Self-supervised Learning for~Class-imbalanced~Image~Data}

\begin{icmlauthorlist}
\icmlauthor{Hye-min Chang}{}
\icmlauthor{~~~~~~~~~~~~~~~~~~~~~Sungkyun Chang}{}
\end{icmlauthorlist}
\center\text{Korea University~~~~~~~~~~~~~~~~~~~~~~~~~~Mimbres Corp.~~~}
% \icmlaffiliation{b}{Mimbres Corp.}
\center\text{}
\vskip 0.3in
]

\begin{abstract}
Class-imbalanced datasets are known to cause the problem of model being biased towards the majority classes. In this project, we set up two research questions: 1) when is the class-imbalance problem more prevalent in self-supervised pre-training? and 2) can offline clustering of feature representations help pre-training on class-imbalanced data? Our experiments investigate the former question by adjusting the degree of {\it class-imbalance} when training the baseline models, namely SimCLR and SimSiam on CIFAR-10 database. 
To answer the latter question, we train each expert model on each subset of the feature clusters. We then distill the knowledge of expert models into a single model, so that we will be able to compare the performance of this model to our baselines.
\end{abstract}

\section{Introduction}
\label{intro}

Self-supervised learning (SSL) is one of the most active topics in the field of computer vision today. Conventional transfer learning scenarios in the visual domain were based on supervised learning, requiring expensive labor to annotate images. SSL has shown that it can overcome such a limitation by providing more general and reusable features to us using unlabeled images. In general, SSL methods\cite{chen2020simple,  chen2021exploring, caron2020unsupervised, caron2021emerging, grill2020bootstrap} consist of two stages: In the pre-training stage, a model is trained by automatic generation of pretext task from the unlabeled images. In the fine-tuning stage, the pre-trained model is transferred to a downstream task using labeled images.

We tackle the problem of {\it class-imbalance} in the pre-training stage of SSL. In the literature\cite{japkowicz2002class, johnson2019survey}, {\it class-imbalance} refers to a classification predictive modeling issue where the number of examples in the training dataset for each class label is not balanced. For example, we can have many training samples of more common classes, while having a few samples for the less common classes. In the case of supervised learning, the imbalanced data distribution  is prone to cause a model biased towards the majority classes. Previously, several techniques based on label information, such as adjustment of class weights for loss function\cite{lin2017focal} and balanced data resampling\cite{buda2018systematic} have been proposed to resolve this problem.  However, the problem of {\it class-imbalance} in SSL has not been discovered well. To further investigate this issue, we set up our research questions:
\begin{questions}
    \item \emph{When does class-imbalance of self-supervised pre-training negatively affect the performance of SimCLR and SimSiam on downstream tasks more?}
    \item \emph{Can offline clustering help SSL make a better pre-training on the class-imbalanced dataset?}
\end{questions}

To outline, we conduct an experiment in Section \ref{Section2} to discover \textbf{RQ1} by impelmenting SSL baslines and generating a class-imbalanced dataset. In Section \ref{Section3}, we verify the idea of \textbf{RQ2} by applying an offline clustering algorithm to one of the baseline SSL models.

\section{Effect of Class-imbalance in SSL}
\label{Section2}
This section investigates the effect of class-imbalance to SSL baseline models. For this, we first generate subsets of CIFAR-10\cite{cifar10} by adjusting the degree of class-imbalance. We then construct two baseline SSL models, and compare the performance of these models trained on differently generated datasets.

\subsection{Generation of Dataset}
We remove different amounts of image samples per class from the original dataset to simulate an imbalanced dataset. Let the number of samples per class in the original balanced dataset be $N_c$ with class label $c = \{0,1,2,...,C-1\}$, where $C=10$ is the number of classes in CIFAR-10. In generation of imbalanced subsets, the number of images per class $\hat{N_c}$ decreases exponentially proportional to a pre-defined  class-imbalance factor $p$ as:
\begin{equation}
    \hat{N_c} = N_c \cdot p^{- c / (C - 1)}.     
\end{equation}

\begin{figure}[t]
\vskip 0.2in
\begin{center}
\centerline{\includegraphics[width=\columnwidth]{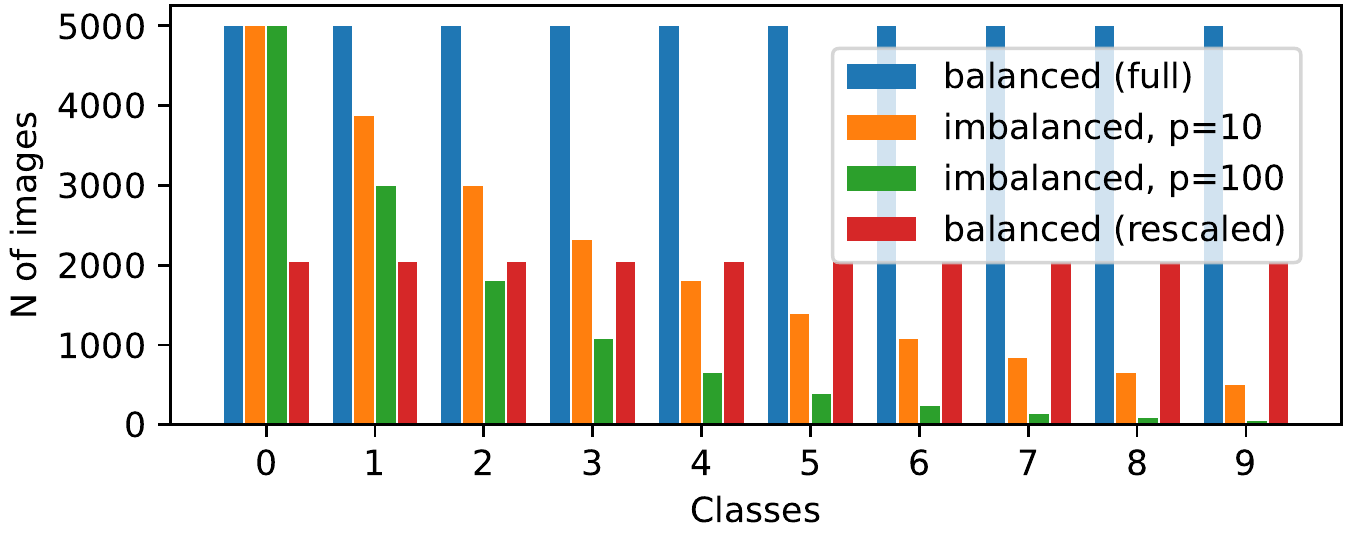}}
\caption{Distribution of the datasets: we denote the original dataset as \textit{balanced (full)}, the generated set with imbalance factor $p$ as \textit{imbalanced}, and the uniformly reduced set as \textit{balanced (rescaled)}.}
\label{fig:dist_gen_data}
\end{center}
\vskip -0.2in
\end{figure}

The distribution of the generated imbalanced subsets are compared with the original dataset in Figure~\ref{fig:dist_gen_data}. For fair evaluation, it is necessary to compare the datasets of the same total number of data. Thus, we also create another balanced set by rescaling the original set.

\subsection{SSL Baselines}
\label{subsec:ssl_baseline}
\begin{figure}[t]
\vskip 0.2in
\begin{center}
\centerline{\includegraphics[width=\columnwidth]{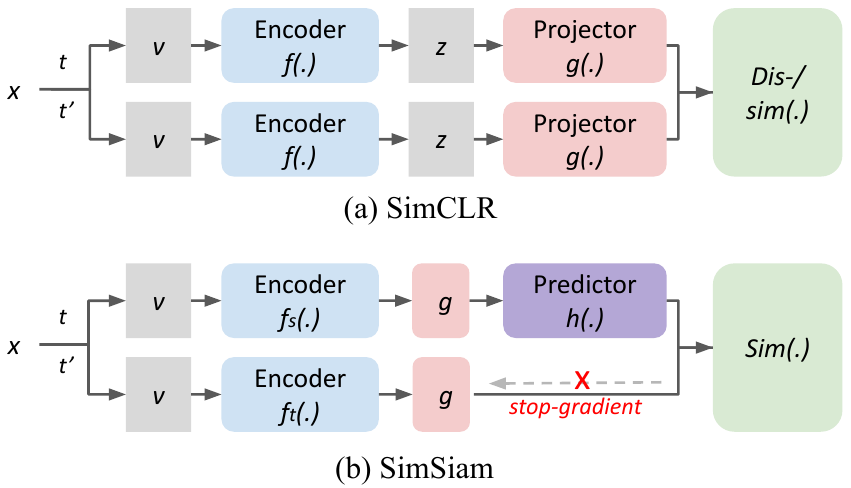}}
\caption{Overview of the SSL baseline models.}
\label{fig:baseline_models}
\end{center}
\vskip -0.2in
\end{figure}

Our SSL baseline models, SimCLR\cite{chen2020simple} and SimSiam\cite{chen2021exploring} are implemented based on the open-source repository\footnote{https://github.com/PatrickHua/SimSiam
}. In Figure~\ref{fig:baseline_models}, each model takes as input a batch that consists of explicitly sampled positive pairs of two different views, where each pair $\{v, v'\}$ is generated from a single image by applying two different augmentations $t$ and $t'$ to an image $x$. These augmentation operations are randomly selected from a set $\mathcal{T}$. As a backbone encoder $f(.)$, we use a variant of ResNet19 architecture following the previous work\cite{chen2021exploring}.

\textbf{SimCLR:} The key feature of our contrastive SSL baseline is that SimCLR maximizes similarity between the positive pairs, and at the same time, dissimilarity between the negative pairs. Here, the negative samples are taken from all other pairs in the batch. In Figure~\ref{fig:baseline_models}(a), $f(.)$ is a shared backbone encoder that outputs a representation $z$. $g(.)$ is a shared MLP projection layer. We will temporarily use $g(.)$ for calculating similarity and dissimilarity of the pairs, and throw $g(.)$ away after finishing pre-training. 

\textbf{SimSiam:} As a non-contrastive baseline, SimSiam is a simplified version of previously proposed BYOL\cite{grill2020bootstrap}. SimSiam maximizes similarity between the positive pairs only. In Figure~\ref{fig:baseline_models}(b), $f_t(.)$ is a teacher network, and $f_s(.)$ is a student network. Here, only a student $f_s(.)$ is connected to an MLP prediction layer $h(.)$. In each step of optimization, the student is forced to mimic the teacher’s representation by maximizing similarity between the output of $h(v)$ and $f(v')_t$. Here, $f_t(.)$ and $f_s(.)$ share the parameters before an update. When updating parameters, we allow only  $h(.)$ and $f_s(.)$ of student to be updated; The teacher's network will not be updated, and it can be implemented by \texttt{torch.detach()} in PyTorch\cite{NEURIPS2019_9015}.

\subsection{Experiment I}
\label{subsec:exp1}
\input{tables/config_exp1}
We shared most of the settings for both SimCLR and SimSiam. Configuration details are listed in Table \ref{tab:exp_setup}. The total number of images in the generated dataset was reduced to less than 40\%, compared to the original CIFAR-10. All models, including pre-trained models on the imbalanced set, were fine-tuned on the balanced set of the same size. To answer RQ1, we attempted to analyze the effect of imbalanced set on pre-training independently from fine-tuning.

\input{tables/results_exp1}
\textbf{Balanced vs. Imbalanced:} Table~\ref{tab:results_exp1} displays the baseline performance evaluated on the test set of CIFAR-10. Overall, “Balanced” always performed better than “Imbalanced”, regardless of other settings. This clearly reveals that the balance between the classes of training data plays a crucial role in self-supervised pre-training stage. More specifically, increasing the imbalance factor $p$ from 10 to 100 consistently degrades the performance of baselines by at least 6\% and up to 20\%. 

\textbf{Imbalanced set is more vulnerable to small data:} 
The original full set always outperformed others. The largest performance gap (31\%) between the "Balanced (full)" and "Imabalanced ($p$=100)" of SimSiam reaffirmed that both class-balance and data size significantly affect the classification accuracy. In particular, when comparing balanced sets, SimSiam showed about 10\% performance improvement when data size increased (12K$\rightarrow$20K). On the other hand, the same rate of data increase in imbalanced set for SimSiam improved the result by 20\%. SimCLR also followed a similar trend to this. Thus, the effect of the amount of data was greater in the imbalanced set. 

\textbf{SimSiam$>$SimCLR:} We observed a clear performance drop of around 20\% for SimCLR compared to SimSiam. This might be unexpected because the performance of our SimCLR (73.57\%) on "Balanced (full)" set underperformed the previous report by nearly 23\%\cite{chen2020simple}. There are two possible explanations for this issue. One is that SimCLR may require much longer training hours than the 300 epochs of our setup. Another performance drop could come from a slightly different backbone encoder, and we also missed the blurring augmentation just to match all the settings with SimSiam. For convenience, the following section will focus on improving the pre-training of the model that reproduced better results---SimSiam.

\section{Offline Clustering on Feature Space}
\label{Section3}
In the previous section, we were able to reproduce the problems caused by class-imbalance in the pre-training stage of SSL by manipulating CIFAR-10 dataset. This section describes the proposed method for alleviating the class-imbalance problem using the information from the distribution of the feature space. Previous works\cite{japkowicz2002class, johnson2019survey} in supervised learning utilized information derived from the labels. Our approach  differs primarily from those for supervised learning in that we do not use the label information directly. 

\subsection{Method}

\begin{figure}[t]
\vskip 0.2in
\begin{center}
\centerline{\includegraphics[scale=0.4]{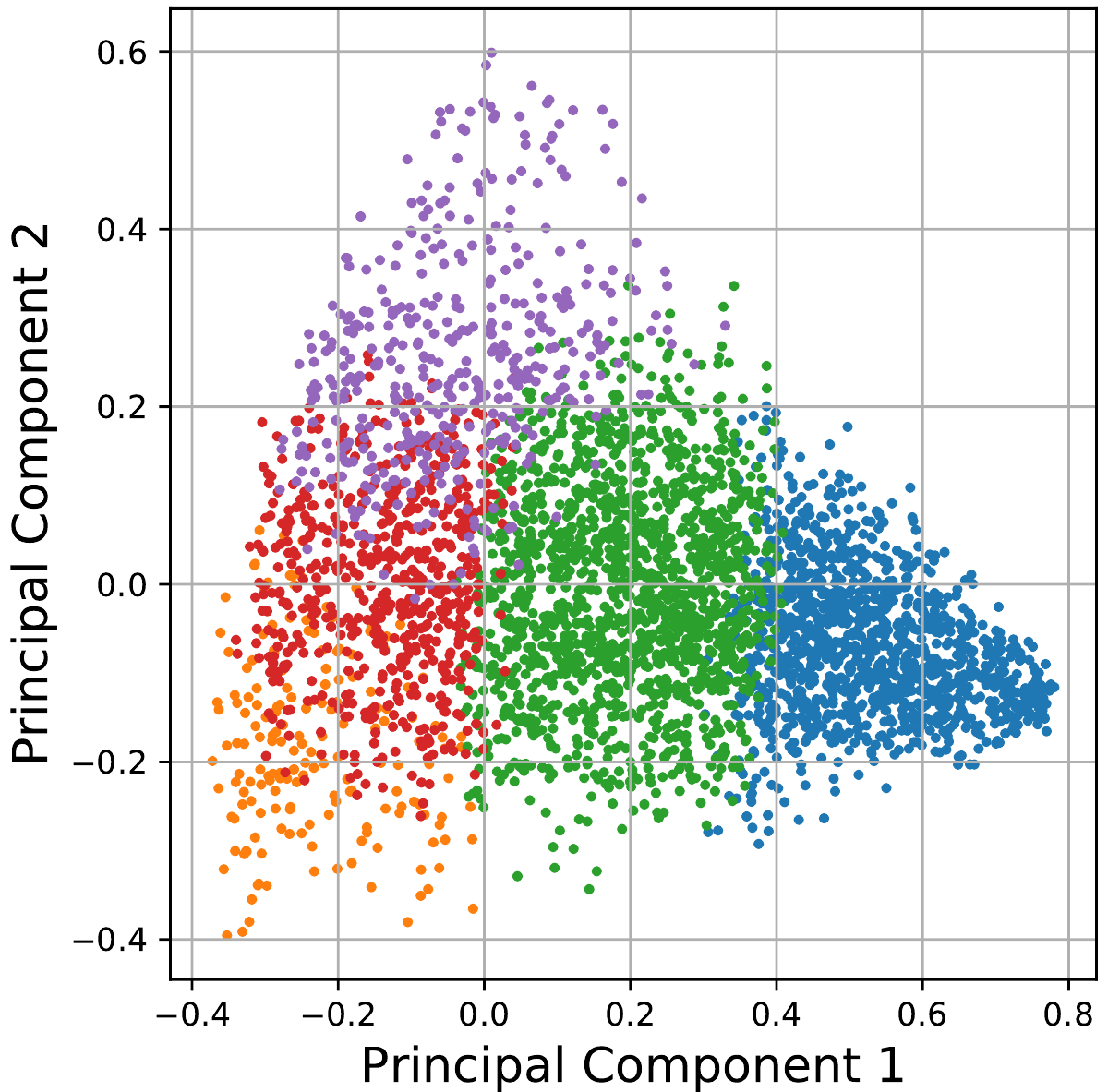}}
\caption{Clusters ($K$=5) of the base model representation of the data points on  imbalanced set ($p$=100). For visualization, 512-$d$ output features of SimSiam\textsubscript{base} trained for 40 epochs are projected onto 2-$d$ space via PCA\cite{abdi2010principal}. }
\label{fig:kmeans}
\end{center}
\vskip -0.2in
\end{figure}
The proposed SSL pre-training is inspired by "Divide and Contrast"\cite{tian2021divide}. The main difference is that we use a non-contrastive SSL model, SimSiam, as a base model. Our pre-training phase consists of four steps: 1) training the base model, 2) dividing the dataset by $k$ feature clusters, 3) training $k$ expert models, and 4) distillation.   
\begin{enumerate}
    \item \textbf{Base Training:} For $M\textsubscript{base}$ epochs, we train a new SimSiam\textsubscript{base} on the whole dataset $\mathbb{D}$ by following the loss function\cite{chen2021exploring}, and the multi-view augmentation and optimizer defined in Table~\ref{tab:exp_setup}.
    \item \textbf{Divide:} We divide the dataset $\mathbb{D}$ into $K$ subsets $\mathbb{D}^{k}$ with $k \in \{1,2,...,K\}$. First, we collect the output representation of SimSiam\textsubscript{base} from $\mathbb{D}$. We then apply \textit{k-means} algorithm to the collected outputs to assign each data point to each cluster as in Figure~\ref{fig:kmeans}.     
    \item \textbf{Expert Training:} For $M\textsubscript{expert}$ epochs, we train each SimSiam\textsubscript{expert(k)} on each dataset $\mathbb{D}^k$ with the loss function, multi-view augmentation, and optimizer defined in Table~\ref{tab:exp_setup}.  
    \item \textbf{Distillation:} For $M\textsubscript{distill}$ epochs, we train a single model SimSiam\textsubscript{student} having the same architecture and parameter as SimSiam\textsubscript{base} by distillation from the SimSiam\textsubscript{base} and SimSiam\textsubscript{expert(k)} with a single-view augmentation. Here the base model is one fixed teacher and the $k$th expert is another teacher that changes upon the assignment of dataset $\mathbb{D}^k$. 
\end{enumerate}
The distillation model's architecture has the same  SimSiam\textsubscript{base} with additional $K+1$ MLP regression heads such that  $r\textsubscript{\{base, 1, 2,...,K\}}(.)$. Here the distillation objective $\mathcal{L}\textsubscript{distill}$ is defined as:
\begin{equation}
    %\begin{gathered}
    \mathcal{L}\textsubscript{distill} = \frac{1}{2}||r\textsubscript{base}(q_s) - q\textsubscript{base}||\hspace{6em}\\
    + \frac{1}{2}||r_k(q_s) - q\textsubscript{expert(k)}||,  
    %\end{gathered}
\end{equation}
where $||.||$ denotes mean squared error (MSE), $q_s$ is the output of SimSiam\textsubscript{student}($v$) with any augmented image $x\in \mathbb{D}$, $q\textsubscript{base}$ is the output of SimSiam\textsubscript{base}($v$) with any augmented image $x\in \mathbb{D}$, and $q\textsubscript{expert(k)}$ is the output of SimSiam\textsubscript{expert(k)}($x_k$) with augmented $x_k\in \mathbb{D}^k$. Note that each $v$ comes from the  single-view augmentation from each image $x$ described in Section~\ref{subsec:ssl_baseline}.

\subsection{Experiment II}
\input{tables/results_exp2}
For fair comparison, we trained every model for 300 epochs. We set $M\textsubscript{base}$, $M\textsubscript{expert}$, and $M\textsubscript{distill}$ as \{40, 180, 80\}, respectively. With this setup, the total number of training epochs for the proposed model is equal to the total number of epochs for the baseline models in Experiment I. For \textit{k-means} clustering, we used \texttt{scikit-learn}\footnote{https://scikit-learn.org/} library, and later replaced it with a GPU-based implementation\footnote{https://github.com/DeMoriarty/TorchPQ}.

Table~\ref{tab:results_exp2} compares the performance of the proposed model with our baselines. Overall, the proposed model denoted as "SimSiam+C+D" outperforms the baselines on the imbalanced sets with $p$=10 and $p$=100. In particular, the improvement for the SimSiam was maximized when $p$=10 by 2.19pp. This revealed that the proposed offline clustering-based pre-training method mitigated the negative effect of class-imbalance dataset.  

To further investigate the effect of distillation alone, we trained a distillation-only model "SimSiam+D" without clustering. This model achieve on a par result with "SimSiam" baseline, and it showed that the effect of distillation alone is quite limited. From this ablation study, it was confirmed that the main factor in performance improvement was the clustering.

\section{Summary and Future Work}
\label{summary}
This project investigated an offline clustering approach to avoid class-imbalance problem in self-supervised pre-training. In Experiment I, we verified that the artificially simulated class-imbalance of training data negatively affected our SSL models in pre-training. Based on the key observation, we employed an unsupervised feature clustering method for dividing the training set into a few subsets. We then trained each expert network on each subset, and finally distilled the base model and expert models into a single network. In Experiment II, 
SimSiam using the proposed training strategy outperformed our baselines on class-imbalanced data. Future directions of this study will include exploring online clustering approaches that can reduce training time, or  searching for more efficient training curriculum. Further investigation of the various types of class-imbalance can be another follow-up topic.
\medskip

\bibliography{main}
\bibliographystyle{icml2021}

\end{document}

%% file: tables/config_exp1.tex
\begin{table}[t!]
  \small
  \caption{Shared configurations for all experiments}
  \label{tab:exp_setup}
  \centering
  \begin{tabular}{ll}
    \toprule
    \textit{Parameter} & \textit{Value} \\
    \midrule
    \textit{(pre-training)} & \\
    Output dimension, $d$ & 512\\
    Batch size & 1,024\\
    Total epochs & 300\\
    Augmentations & \cite{caron2020unsupervised}\\
    Learning rate schedule & Cosine\\
    SimCLR.optimizer & LARS\cite{you2017large}\\
    SimCLR.weight decay & 1e-6\\
    SimCLR.base learning rate & 0.3\\
    SimSiam.optimizer & SGD\\
    SimSiam.weight decay & 1e-5\\
    SimSiam.base learning rate & 0.03\\
    \midrule
    \textit{(fine-tuning)} & \\
    Batch size & 256\\
    Total epochs & 100\\
    Augmentations & \cite{caron2020unsupervised}\\
    Optimizer & SGD \\
    Learning rate & 30.0\\
    Momentum & 0.9\\
    \midrule
    \textit{(dataset)} & \\
    Image size & 32 $\times$ 32\\
    Class imbalance factor, $p$ & $[10, 100]$\\
    N of images for train (full) & 50,000\\
    N of images for train (p=10) & 20,431\\
    N of images for train (p=100) & 12,409\\
    N of images for test & 10,000\\
  \bottomrule
  \end{tabular}
\end{table}

%% file: tables/results_exp1.tex
\begin{table}[t!]
  \footnotesize %small
  \caption{Classification accuracy (\%) of the baseline SSL models on the "imbalanced" vs. "balanced" sets in Experiment I: $N$ is the total number of training examples. "rs." represents a randomly rescaled, fixed dataset. The accuracy in the blanks is from the full epoch experiment reported in \cite{chen2020simple, chen2021exploring}.}
  \label{tab:results_exp1}
  \centering
  \begin{tabular}{llll}
    \toprule
    \textit{CIFAR-10 Subset} & \textit{N} & \textit{SimCLR} & \textit{SimSiam}\\
    \midrule
    Imbalanced (p=10) & 20K & 66.52 & 81.16\\
    Imbalanced (p=100) & 12K & 62.18 & 68.54\\
    Balanced (rs.20K) & 20K & 71.25 & 87.14\\
    Balanced (rs.12K) & 12K & 69.82 & 79.23\\
    \midrule
    Balanced (full) & 50K & 73.57 (91.1) & 90.40 (91.18)\\
  \bottomrule
  \end{tabular}
\end{table}

%% file: tables/results_exp2.tex
\begin{table}[t!]
  \small
  \caption{Classification accuracy (\%) of SSL models on the imbalanced vs. balanced sets in Experiment II: "+C" indicates clustering, and "+D" indicates distillation from multiple experts.\\}
  \label{tab:results_exp2}
  \centering
\begin{tabular}{@{}lrrrr@{}}
\toprule
\multirow{2}{*}{\textit{\begin{tabular}[c]{@{}l@{}}CIFAR-10\\ Subset\end{tabular}}} & \multicolumn{2}{c}{\textit{Baseline}} & \multicolumn{1}{l}{\multirow{2}{*}{\textit{\begin{tabular}[c]{@{}l@{}}SimSiam\\ +C+D\end{tabular}}}} & \multicolumn{1}{l}{\multirow{2}{*}{\textit{\begin{tabular}[c]{@{}l@{}}SimSiam\\ +D\end{tabular}}}} \\ \cmidrule(lr){2-3}
 & \multicolumn{1}{l}{\textit{SimCLR}} & \multicolumn{1}{l}{\textit{SimSiam}} & \multicolumn{1}{l}{} & \multicolumn{1}{l}{} \\ \midrule
Imb. (p=10) & 66.52 & 81.16 & \textbf{83.35} & 80.98 \\
Imb. (p=100) & 62.18 & 68.54 & \textbf{69.62} & 68.21 \\
Bal. (rs.20K) & 71.25 & 87.14 & n/a & n/a \\
Bal. (rs.12K) & 69.82 & 79.23 & n/a & n/a \\ \bottomrule
\end{tabular}
\end{table}